\documentclass[letterpaper, 10 pt, journal, twoside]{IEEEtran}
\usepackage{comment}
\usepackage{siunitx}
\usepackage{relsize}
\usepackage{ifthen}
\usepackage[colorinlistoftodos]{todonotes}

\usepackage{graphics} %
\usepackage{rotating}
\usepackage{color}
\usepackage{enumerate}
\usepackage[T1]{fontenc}
\usepackage{psfrag}
\usepackage{epsfig} %
\usepackage{booktabs}
\usepackage{graphicx,url}
\usepackage{multirow}
\usepackage{array}
\usepackage{latexsym}
\usepackage{amsfonts}
\usepackage{amsmath}
\usepackage{amssymb}
\usepackage{xstring}
\usepackage{multirow}
\usepackage{xcolor}
\usepackage{prettyref}
\usepackage{flexisym}
\usepackage{bigdelim}
\usepackage{breqn} %
\usepackage{listings}
\let\labelindent\relax
\usepackage{enumitem}
\usepackage{xspace}
\usepackage{bm}
\graphicspath{{./figures/}}
\usepackage{tikz}
\usetikzlibrary{matrix,calc}

\usepackage{mdwlist}

\let\itemize\undefined
\makecompactlist{itemize}{stditemize}

\usepackage{etex}
\usepackage{algorithm, algorithmic}
\usepackage{subcaption}
\usepackage{amsmath}
\usepackage{lipsum}

\usepackage{tabularx, colortbl}

\usepackage{filecontents}

\DeclareCaptionLabelSeparator{periodspace}{.\quad}
\IEEEoverridecommandlockouts
\usepackage{cite}
\usepackage{multicol}
\usepackage[bookmarks=true]{hyperref}

\usepackage{amsthm}
\newtheoremstyle{mystyle}%
  {}%
  {}%
  {\itshape}%
  {}%
  {\bfseries}%
  {.}%
  { }%
  {\thmname{#1}\thmnumber{ #2}\thmnote{ (#3)}}%
\theoremstyle{mystyle}
\newrefformat{prob}{Problem\,\ref{#1}}
\newrefformat{def}{Definition\,\ref{#1}}
\newrefformat{sec}{Section\,\ref{#1}}
\newrefformat{sub}{Section\,\ref{#1}}
\newrefformat{prop}{Proposition\,\ref{#1}}
\newrefformat{app}{Appendix\,\ref{#1}}
\newrefformat{alg}{Algorithm\,\ref{#1}}
\newrefformat{cor}{Corollary\,\ref{#1}}
\newrefformat{thm}{Theorem\,\ref{#1}}
\newrefformat{lem}{Lemma\,\ref{#1}}
\newrefformat{fig}{Fig.\,\ref{#1}}
\newrefformat{tab}{Table\,\ref{#1}}

\newcommand{\bdmath}{\begin{dmath}}
\newcommand{\edmath}{\end{dmath}}
\newcommand{\beq}{\begin{equation}}
\newcommand{\eeq}{\end{equation}}
\newcommand{\bdm}{\begin{displaymath}}
\newcommand{\edm}{\end{displaymath}}
\newcommand{\bea}{\begin{eqnarray}}
\newcommand{\eea}{\end{eqnarray}}
\newcommand{\beal}{\beq \begin{array}{ll}}
\newcommand{\eeal}{\end{array} \eeq}
\newcommand{\beas}{\begin{eqnarray*}}
\newcommand{\eeas}{\end{eqnarray*}}
\newcommand{\ba}{\begin{array}}
\newcommand{\ea}{\end{array}}
\newcommand{\bit}{\begin{itemize}}
\newcommand{\eit}{\end{itemize}}
\newcommand{\ben}{\begin{enumerate}}
\newcommand{\een}{\end{enumerate}}

\newcommand{\eg}{\emph{e.g.,}\xspace}
\newcommand{\ie}{\emph{i.e.,}\xspace}
\newcommand{\myParagraph}[1]{{\bf #1.}\xspace}
\renewcommand{\boldsymbol}[1]{{\bm #1}}

\newcommand{\hide}[1]{}

\newcommand{\hiddenText}{{\color{gray} hidden text.}}
\newcommand{\hideWithText}[1]{\hiddenText}

\newcommand{\SLfour}{\ensuremath{\mathrm{SL}(4)}\xspace}

\newcommand{\blue}[1]{{\color{blue}#1}}

\newcommand{\linkToPdf}[1]{\href{#1}{\blue{(pdf)}}}
\newcommand{\linkToPpt}[1]{\href{#1}{\blue{(ppt)}}}
\newcommand{\linkToCode}[1]{\href{#1}{\blue{(code)}}}
\newcommand{\linkToWeb}[1]{\href{#1}{\blue{(web)}}}
\newcommand{\linkToVideo}[1]{\href{#1}{\blue{(video)}}}
\newcommand{\linkToMedia}[1]{\href{#1}{\blue{(media)}}}
\newcommand{\award}[1]{\xspace} %

\usepackage{float}
\usepackage{hyperref}
\usepackage{graphicx}
\usepackage{epstopdf}
\usepackage{epsfig}
\usepackage{amsmath}
\usepackage[capitalise]{cleveref}
\usepackage[normalem]{ulem}
\usepackage{adjustbox}

\newcommand{\name}{FOUND-IT\xspace}

\newcommand{\vgstwo}{VGGT-SLAM 2.0\xspace}

\newcommand{\thres}{threshold}
\newcommand{\batch}{batch}
\newcommand{\sg}{SemanticGS}
\newcommand{\ogs}{OpenGS}

\newcommand{\os}{open-set\xspace}

\newcommand{\tileSize}{$n$}

\DeclareRobustCommand{\IEEEauthorrefmark}[1]{\smash{\textsuperscript{\footnotesize #1}}}

\newcolumntype{C}{>{\centering\arraybackslash}X} 

\usepackage{xr}

\title{\LARGE \bf{FOUND-IT: Foundation-model-first Task-driven \\  3D Scene Graphs with Granularity on Demand}}

\author{Dominic Maggio$^*$\IEEEauthorrefmark{1},
                  Nicolas Gorlo$^*$\IEEEauthorrefmark{1},
                  Kris Hauser\IEEEauthorrefmark{2},
                  Luca Carlone$^\dagger$\IEEEauthorrefmark{1}
    \thanks{$^1$Laboratory for Information \& Decision Systems, Massachusetts Institute of Technology 
    Cambridge, MA, USA. Email: \{drmaggio, ngorlo, lcarlone\}@mit.edu.}
    \thanks{$^2$Samsung Research America}
    \thanks{$^\dagger$Luca holds concurrent appointments as a faculty at the Massachusetts Institute of Technology and as an Amazon Scholar. This paper describes work performed at MIT and is not associated with Amazon.}
    \thanks{This project was partially funded by Samsung Research America and  
    by the NSF Graduate Research Fellowship Program under Grant 2141064. }
   \thanks{$^*$equal contribution.}}

\begin{document}
\bstctlcite{IEEEexample:BSTcontrol}

\maketitle
\thispagestyle{empty}
\pagestyle{empty}

\begin{abstract}
    We present the first approach to build hierarchical task-driven 3D scene graphs of arbitrary indoor or outdoor environments 
    using an uncalibrated monocular camera in real-time. We leverage geometric foundation models to estimate geometric attributes 
    of the scene graph (\eg object bounding boxes), but we also observe that traversability 
    information (the “places” layer of a scene graph) can be directly reconstructed by adding an extra head to 
    existing geometric foundation models, like VGGT. Our approach is task-driven in the sense that we adjust 
    the granularity of the objects and regions in the map depending on the task; for instance, during a manipulation task, 
    our approach is able to resolve small knobs on a stove, while during a navigation task it can focus on 
    large objects (\eg the entire stove). However, in a major departure from related work, we consider the realistic case where 
    the list of tasks is not predefined and fixed, but evolves as the robot operates. This naturally allows 
    dealing with complex loco-manipulation tasks, where the robot can dynamically adjust its representation as the task unfolds. 
    We dub the resulting approach \name. \name also includes an agentic approach to query information in the 
    scene graph. %
    In addition to achieving 79\% higher accuracy on the ASHiTA SG3D task grounding benchmark, we demonstrate \name runs in real-time 
    on a ground robot using a Jetson Thor. Furthermore, to highlight the robustness of our method, we demonstrate constructing 3D scene 
    graphs on casually captured realtor apartment tours from YouTube. Code will be made available upon publication.
\end{abstract} %

\section{Introduction}
\label{sec:intro}

Actionable and versatile 3D scene understanding in complex indoor and outdoor environments 
is a fundamental step toward spatial intelligence in robotics. 
Towards this goal, \emph{3D scene graphs}~\cite{Armeni19iccv-3DsceneGraphs, Wu21cvpr-SceneGraphFusion, Hughes24ijrr-hydraFoundations, Maggio24ral-clio, Schmid24rss-khronos} 
construct hierarchical metric-semantic models of a scene by creating a map of objects, places (a topological representation of traversable space), and regions (such as rooms), among other abstractions. 
However, two core limitations of 3D scene graph construction methods are their \emph{reliance on depth sensing} and 
the difficulty in ensuring they describe 
concepts at the correct granularity to support a robot during its mission. 
Current real-time 3D scene graphs depend on calibrated stereo cameras (or RGB-D sensors) and complex pipelines which limits their 
ability to be easily deployed, maintained, or adapted. 
These approaches form objects at mapping time and must correctly track and associate objects over time. 
Additionally, representation of places and 
regions in many 3D scene graphs~\cite{Maggio24ral-clio, Werby24rss-hovsg, Hughes24ijrr-hydraFoundations} 
rely on purely geometric approaches that implicitly assume structured, indoor rooms. 
While some scene graph frameworks~\cite{Ray24isrr-tamp,Strader24ral-automaticAbstractions,Gorlo26cvpr-DAAAM} 
have expanded to both indoor and outdoor settings, 
they universally depend on complex architectures and calibrated stereo cameras or depth sensors which limit their accessibility and deployment. 

\begin{figure}[t]
    \centering
    \includegraphics[width=.48\textwidth]{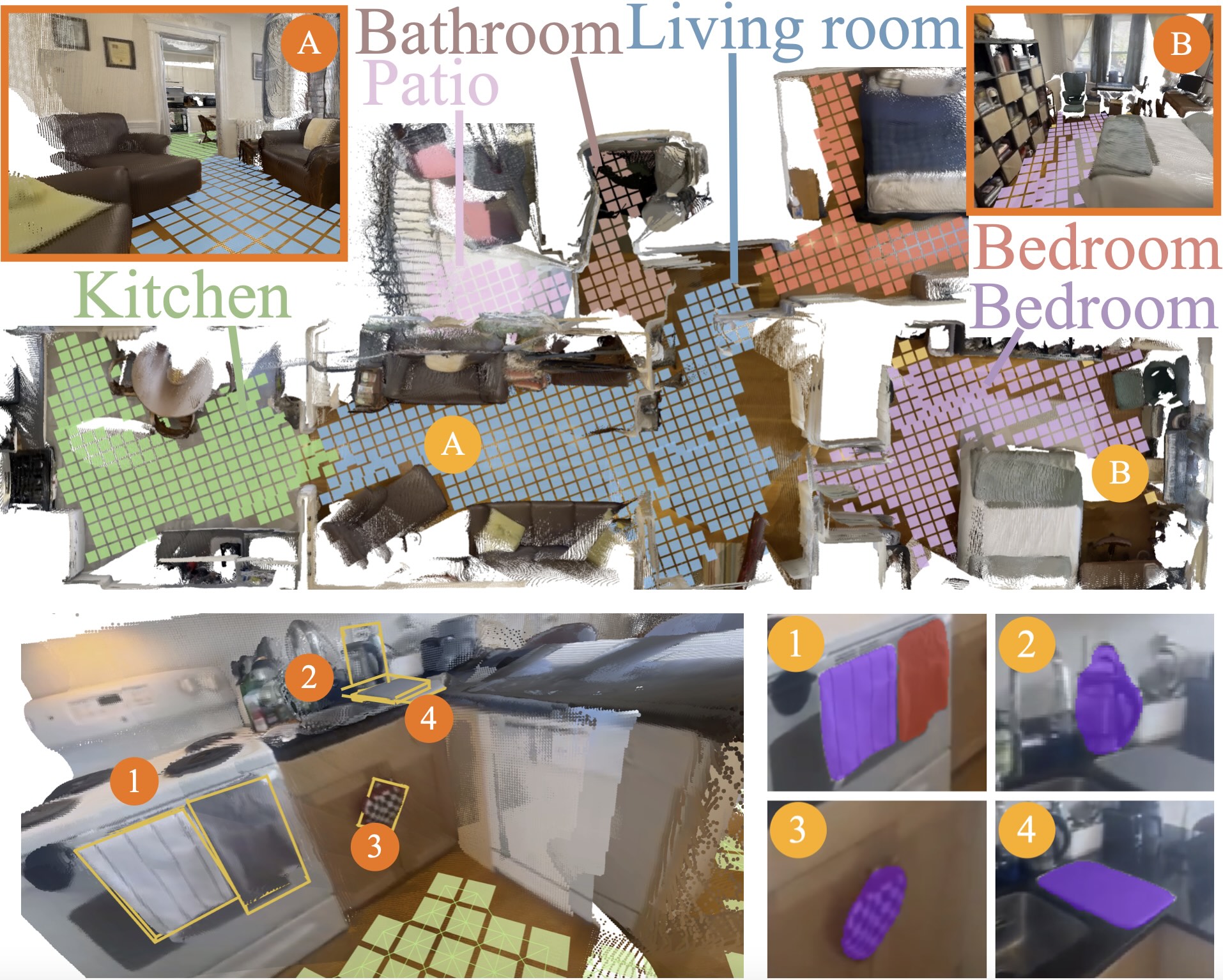}
    \caption{Example of scene graphs constructed by \name using home tours videos from YouTube. Top: example 
    room detections. The places layer, used for planning over traversable space, 
    is shown as tiles on the floor, and clustered into regions. Bottom: example object detections for the queries (1) oven towel, (2) water kettle, (3) oven mitt, and 
    (4) cutting board.}
    \label{fig:intro}
\end{figure}

Conversely, recent advances in \emph{Geometric Foundation Models} (GFMs)~\cite{Wang25cvpr-vggt,Leroy24eccv-mast3r,
Wang24cvpr-dust3r,Chen25arxiv-ttt3r,Maggio26rss-VGGT-SLAM2} have revolutionized 3D reconstruction and SLAM by enabling accurate, dense 3D mapping via greatly simplified architectures and uncalibrated monocular cameras. 
Despite GFMs turning geometric reconstruction into a plug-and-play algorithm, 
limited work has explored extending GFMs to construct 3D scene graphs, with some recent work extending GFMs for short-sequence 
visual question answering~\cite{Fan25arxiv-vlm3r, Wu25arxiv-spatialmllm} and semantic grounding~\cite{Koch25arxiv-unite}.

In addition to their reliance on complex pipelines, current approaches are fundamentally limited by their semantic expressiveness. 
Many real-time 3D scene graph systems~\cite{Armeni19iccv-3DsceneGraphs, Wu21cvpr-SceneGraphFusion, Hughes24ijrr-hydraFoundations} 
either rely on closed-set segmentation models, which limits the variety of concepts they can capture, 
or leverage open-set vision-language models such as~\cite{Radford21icml-clip,Zhai23iccv-siglip} 
by arbitrarily fixing granularity of 
semantics at mapping 
time~\cite{Gorlo26cvpr-DAAAM, Jatavallabhula23rss-ConceptFusion, Takmaz23neurips-openmask3d, Gu24icra-conceptgraphs, Werby24rss-hovsg, Kassab24arxiv-BareNecessities}.
To effectively leverage the expressiveness of open-set models, a 3D map must resolve objects at the \emph{correct granularity},
avoiding creating objects that are too coarse (which discards relevant information) or too fine (which misses 
higher-level semantic concepts and potentially incurs high memory use). 
For instance, a home robot tasked with even a simple objective like turning off the stove must be able to adaptively 
represent the same area of a scene with course 
semantic concepts (the entire stove) to navigate to the correct location and finer-concepts ---such as individual knobs--- 
after reaching the stove, to 
execute specific manipulation tasks.

As an initial step to address this issue of granularity, prior work~\cite{Maggio24ral-clio, Maggio25arxiv-BayesianFields, Chang25cvpr-ashita} 
created task-driven approaches where the map is formed at a granularity to support a specific (pre-defined) set of natural language tasks.
However, usability is significantly constrained because \emph{the list of tasks must be 
predefined} and cannot be updated dynamically during or after mapping. 
In practice, a robot needs a representation that can  support changing granularity as it executes its objectives. 

\myParagraph{Contributions} We propose \name, a real-time, open-set, hierarchical 3D scene graph construction system, 
which generates maps in indoor and outdoor environments and dynamically adjusts their granularity depending on the task.
Our approach requires only monocular images as input and allows for task-driven, open-set mapping of 
objects and regions without requiring a fixed a-priori list of tasks (intuitively, the approach is \emph{prompted} with a task at runtime, just like a Vision-Language-Action model). 
Through tightly coupled integration with 
Geometric Foundation Models, we build \name on top \vgstwo~\cite{Maggio26rss-VGGT-SLAM2},
creating the first 3D scene graph construction method using an uncalibrated monocular camera.

Our first contribution (\cref{sec:objects}) is a simplified yet powerful approach for creating a 3D scene graph with a task-driven object-granularity at query time without requiring a pre-defined task list.
Rather than defining objects as incoming frames are being processed, we create a visual memory 
layer which maintains semantic information through keyframe-wise semantic embeddings and allows for real-time, open-set 3D object querying and mapping. Upon querying, objects are stored explicitly in the map's cache memory. 

Our second contribution (\cref{sec:places}) is a 3D topological graph representing traversable space (referred to as the \emph{places layer}) which maintains simplicity 
and generalizability by using a novel ground segmentation head appended to the GFM. This allows us to map traversable regions of diverse scenes ranging from indoor to outdoor environments.
By studying the visual attention layers of a GFM we show the best layers for ground segmentation are intermediate network layers. 
We further show that our segmentation approach generalizes to multiple GFMs.

Our third contribution (\cref{sec:regions}) is an efficient query-time clustering algorithm that groups places into open-set regions (\eg rooms in indoor environments), determining both spatial extent and semantic granularity based on the query.

Our fourth contribution (\cref{sec:agent}) is to integrate the algorithms above into an agentic system that builds a task-oriented 3D scene graph by querying for objects related to a task. Unlike other agentic 3D spatial memory systems~\cite{Anwar25icra-remembr,Gorlo26cvpr-DAAAM} that use LLM agents to retrieve from a pre-built representation, our agent incrementally builds a 3D scene graph as queries come in, constructing the right granularity for each object at query time.

Our fifth contribution (\cref{sec:experiments}) is a suite of experiments, which ---in addition to demonstrating top performance on open-set 3D object detection 
on the Clio dataset~\cite{Maggio24ral-clio} and a 79\% improvement on the 
ASHiTA SG3D benchmark~\cite{Chang25cvpr-ashita,Zhang24arxiv-taskOrientedGrounding}--- shows that \name runs in real-time 
onboard a ground robot. Furthermore, to highlight that our system is generalizable, we 
demonstrate constructing 3D scene graphs on in-the-wild cellphone images from realtor apartment walk-through tours 
from the internet (\cref{fig:intro}).  %

\section{Related Work}
\label{sec:related_works}

\myParagraph{Metric-Semantic SLAM and 3D Scene Graphs}
Real-time metric-semantic SLAM and 3D scene-graph construction are a popular framework for spatial memory in robotics.
Metric-semantic SLAM systems~\cite{Rosinol21ijrr-Kimera,McCormac17icra-semanticFusion,Narita19iros-metricSemantic,Grinvald19ral-voxbloxpp,Schmid24rss-khronos} ground discrete semantic information (\eg object class/instance) in 3D maps for object-centric scene understanding. 3D Scene graphs~\cite{Armeni19iccv-3DsceneGraphs,Wu21cvpr-SceneGraphFusion,Hughes24ijrr-hydraFoundations} structure these outputs into graphs with semantic and relational information for downstream reasoning.
More recent open-vocabulary 3D systems~\cite{Jatavallabhula23rss-ConceptFusion,Takmaz23neurips-openmask3d,Gu24icra-conceptgraphs,Koch24cvpr-Open3DSG,Werby24rss-hovsg,Kassab24icra-lexis,Kassab24arxiv-BareNecessities,Gorlo26cvpr-DAAAM} lift open-vocabulary semantic annotations (language-image embeddings or language descriptions) into 3D, but inherit the granularity of their underlying 2D segmentation models, and likewise set the granularity of regions at mapping time.
Task-driven mapping~\cite{Agia22corl-Taskography,Maggio24ral-clio,Maggio25arxiv-BayesianFields,Chang25cvpr-ashita} addresses the granularity issue by conditioning cluster formation on a predefined list of tasks at mapping time.
These systems often assume posed RGB-D data as sensor input and pre-cluster semantic features at mapping time (committing to a granularity a priori), or ---in the latter case of task-driven systems--- commit to a fixed list of tasks a priori.
\name removes the pre-defined task list by storing a keyframe-based visual memory and determining granularity at query time. It further takes uncalibrated monocular RGB video-stream as input, and extracts objects, places, and regions using a foundation-model-first approach.

\myParagraph{Geometric Foundation Models (GFMs) for 3D Scene Understanding}
Feed-forward GFMs regress 3D structure from uncalibrated images in a single pass~\cite{Wang24cvpr-dust3r,Leroy24eccv-mast3r,Wang25cvpr-vggt,Wang25cvpr-cut3r,Wang25iclr-pi3,Chen25arxiv-ttt3r,Keetha25arxiv-mapanything}, achieving accurate, dense reconstruction from in-the-wild videos.
Recent SLAM systems extend these feed-forward predictors to incremental dense SLAM over long trajectories~\cite{Murai25cvpr-mast3rslam,Deng25arxiv-vggtlong,Maggio25neurips-VGGT-SLAM,Maggio26rss-VGGT-SLAM2}.
A few works couple GFM features with a VLM for short-sequence 3D question answering~\cite{Fan25arxiv-vlm3r,Wu25arxiv-spatialmllm} and 
object semantics~\cite{Koch25arxiv-unite} but do not scale to long-horizon roll-outs nor build hierarchical models.
\name, our real-time 3D scene graph construction system, leverages GFMs for building a hierarchical scene graph based on task-queries 
and estimates traversability directly from intermediate GFM tokens.

\myParagraph{LLM-Agent Spatial Memory}
Recent agentic systems~\cite{Rana23corl-sayplan,Anwar25icra-remembr,Xie24arxiv-EmbodiedRAG,Saxena25corl-GraphEQA,Honerkamp24ral-MoMa,Gorlo26cvpr-DAAAM} use LLM agents that iteratively call retrieval or planning tools over a 3D scene graph or spatial memory that is built independently of the query. SG-Nav~\cite{Yin24neurips-sgnav} additionally lets the agent actively direct exploration, but the underlying representation granularity stays fixed.
By using \name , an LLM agent instead actively constructs a query-centric 3D scene graph and determines what to store in the scene graph representation based on queries.

\section{\name: Feed-forward-based Task-Driven Open-Set 3D Scene Graphs}
\label{sec:method}

In this section, we present our design of \name, which takes in uncalibrated monocular images 
and forms a hierarchical 3D scene graph that supports \os object querying, \os room quering, path planning, 
and can be orchestrated by an agent for visual question answering. 
In \cref{sec:geometric}, we provide an overview of the 3D geometric 
mapping framework used by \name. In \cref{sec:objects}, we describe our object layer. 
In \cref{sec:places}, we present our approach for extracting traversable space, \ie the places layer.
Based on text queries, this layer can then be clustered into task-relevant regions (\cref{sec:regions}).
Finally, we describe how we integrate the system with agentic reasoning in \cref{sec:agent}.

\subsection{Geometric Mapping}
\label{sec:geometric}

We follow the approach of \vgstwo~\cite{Maggio26rss-VGGT-SLAM2} to create a geometric map which consists of smaller submaps produced from
a GFM. Given a stream of images, keyframes are designated based on disparity, and once a fixed-size batch of keyframes is collected,
they are passed to a GFM to obtain dense depth maps, depth confidence maps, poses, and camera intrinsics. In \cref{sec:places}, we describe how we additionally
add a ground mask segmentation output, which is used to build the places layer.
Submaps are chained together, and global optimization is performed to support loop closures using a
factor graph optimized on the \SLfour manifold.
We use VGGT~\cite{Wang25cvpr-vggt} as our default GFM and 
demonstrate our entire pipeline (including places segmentation), %
can be easily adapted to different foundation models by also using two variants of 
Depth Anything 3~\cite{Lin26iclr-depthAnything} as drop-in replacements (\cref{sec:experiments}). 

Since actively storing all dense 3D points is memory intensive, we optionally deploy a 
sparsification approach to downsample points in a submap by voxelization. In \cref{sec:objects}, we demonstrate 
\name can keep the overall map sparse, while densely representing queried objects. 

\subsection{Objects Layer}
\label{sec:objects}

A significant challenge in building a task-driven scene graph is that the granularity required from the mapping system depends on the agent's 
objectives, yet in practice these tasks may not be known at mapping time and might evolve as the robot operates. 
Still, per-frame object-extraction pipelines~\cite{Maggio24ral-clio, Gu24icra-conceptgraphs, Maggio25arxiv-BayesianFields} form and commit to object instances as frames stream in, fixing granularity before the query is known (or assume the query/task is known a-priori). 
The object layer of \name instead handles instance formation at query-time by constructing a two-stage memory: a \emph{visual memory} retains keyframes along with their corresponding semantic embedding vector, and a \emph{cached memory} contains extracted objects with their corresponding point cloud and 3D oriented bounding box.

\myParagraph{Visual Memory} To defer explicit object formation, we keep a keyframe-indexed embedding rather than pre-extracted masks or crops. When constructing each submap, all keyframes are passed in batch to a CLIP visual encoder (Perception Encoder~\cite{Bolya25neurips-perceptionEncoder}) to obtain a semantic embedding per keyframe. Note that keyframes 
are already stored on disk by \vgstwo for potential future loop closures. 
Embedding whole images rather than mask crops (\eg as in ~\cite{Maggio24ral-clio, Gu24icra-conceptgraphs, Maggio25arxiv-BayesianFields}) (i) keeps the granularity open to be decided after mapping, (ii) gives the VLM more semantic context, (iii) reduces compute to one embedding per keyframe rather than one embedding per mask, and (iv) removes the need for 3D object tracking and data association.
When provided with task context in the form of a text query,
we compute the CLIP embedding of the query (which we can batch in the event of multiple prompts) and 
compute the cosine score between the text embedding and keframe embeddings. 
Our objective now is to use the cosine scores to identify images to pass to SAM3~\cite{Carion25arxiv-sam3} to obtain 2D segmentation masks.
Unlike \vgstwo, which passes only the top-scoring keyframe to SAM3 and therefore cannot recover multiple object instances, 
we continue passing keyframes to SAM3 as long as their camera frustum does not already view a mapped object of the current query and SAM3 identifies objects. 
The resulting 2D masks are back-projected to 3D using the corresponding depth, intrinsics, and confidence mask. 
Since depth images are stored on disk, the overall map can stay sparse while only queried objects become dense. 
A full object query runs in 100ms, $4\times$ faster than \vgstwo's object query, on a 3090 GPU.

\myParagraph{Cached Memory} Once a query resolves an object, 
it is stored in the cache as a point cloud with a 3D oriented bounding box.
Subsequent queries first search the cache and fall back to the visual memory if the query is not present.
Intuitively, this cache not only saves computation time by eliminating recomputation for repeated queries, 
but also allows us to form a flexible task-driven scene graph incrementally as the agent performs its tasks.

\subsection{Places Layer}
\label{sec:places}

The places layer has two goals: (i) incrementally map traversable areas to enable planning through the 
scene (\eg between the robot's current pose and an object of interest) while being deformable in the event of loop closures and 
(ii) serve as primitives that can be 
clustered into higher-level semantic constructs (\ie regions in \cref{sec:regions}). 
We aim for a simple design of the places layer that works across diverse indoor and outdoor environments. 
Strictly geometric free-space clustering approaches add complexity and often fail to generalize between indoor and outdoor environments, 
so we instead fine-tune the output of the GFM to predict traversable ground and fit tiles on the ground to form 
the places layer.

\begin{figure}[h]
    \centering
    \includegraphics[width=.49\textwidth]{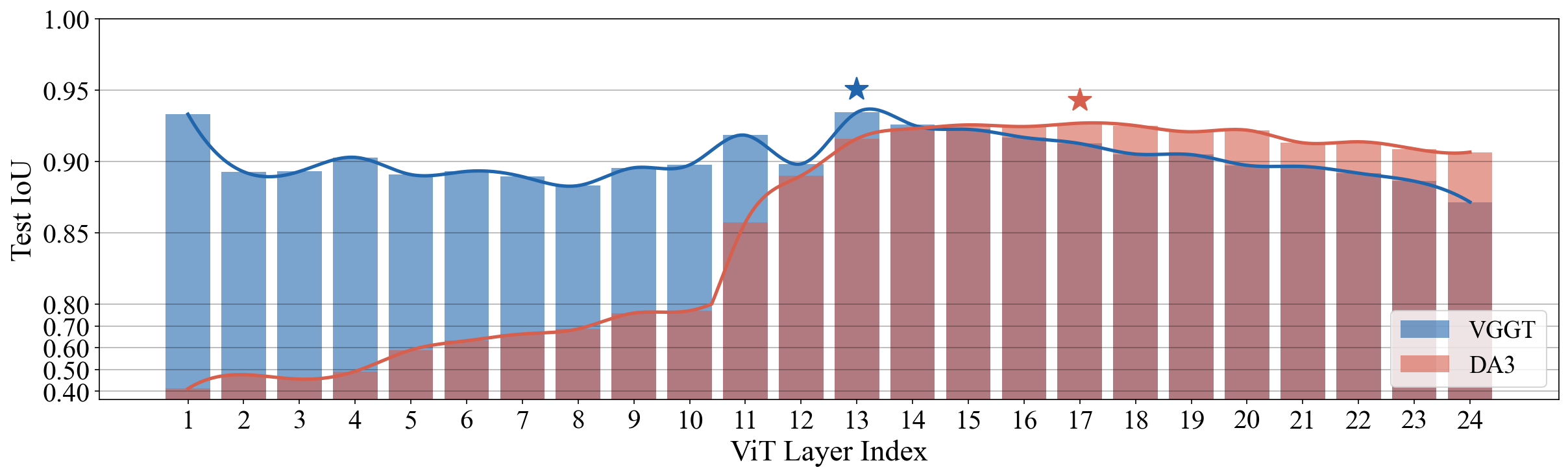}\vspace{-2mm}
    \caption{Ground segmentation performance from training a new head on tokens for each layer 
    of VGGT and Depth Anything 3 model DA3-LARGE-1.1. A star denotes the best layer.}
    \vspace{-2mm}
    \label{fig:places_layers}
\end{figure}

\begin{figure}[h]
    \centering
    \vspace{-1mm}
    \includegraphics[width=0.489\textwidth]{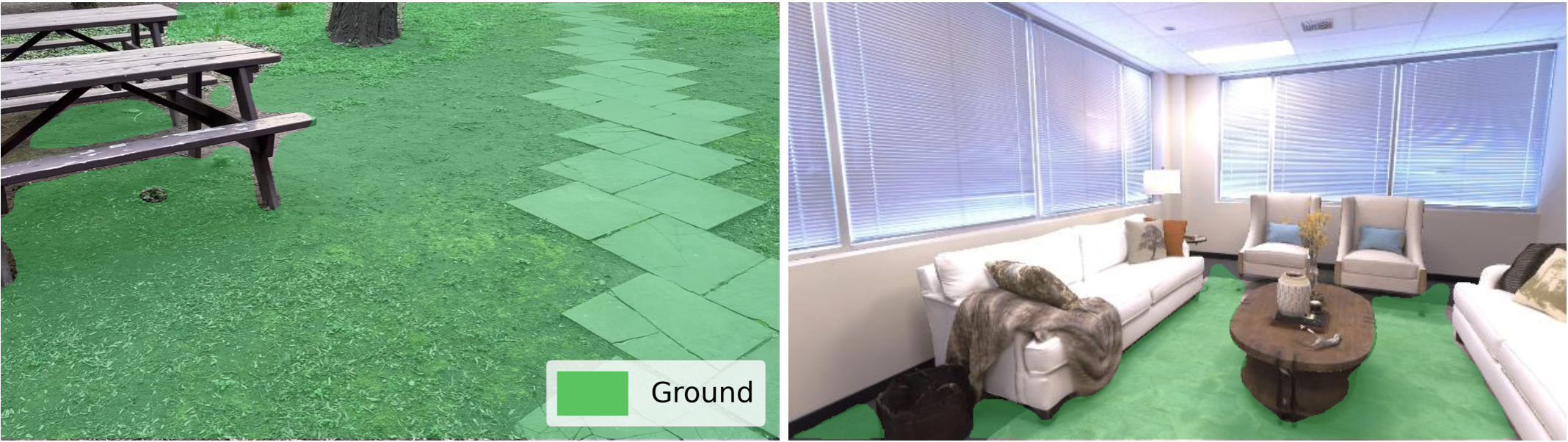}\\
    \caption{Example of estimated ground segmentation on novel outdoor (left) and indoor (right) scenes using our 
    VGGT ground segmentation head.}
    \vspace{-1mm}
    \label{fig:ground_seg}
\end{figure}

\myParagraph{Ground Segmentation}
To produce ground segmentation of keyframes, we train a convolutional decoder head that takes image tokens from the GFM as input.
While a potential choice is to use final-layer tokens, we empirically find earlier layers outperform the final layer for 
ground segmentation. 
Orthogonally, Perception Encoder~\cite{Bolya25neurips-perceptionEncoder} also 
observes that select intermediate layers of a vision-language model are better 
suited for training downstream tasks. 
In~\cref{fig:places_layers}, we train a segmentation head on tokens from each of the 24 layers of Depth Anything~\cite{Lin26iclr-depthAnything} DA3-LARGE-1.1 
and VGGT~\cite{Wang25cvpr-vggt} and by showing test IoU on novel scenes, we empirically identify which intermediate 
layers are well suited for the ground segmentation task. 
Using the optimal layer of tokens from the GFM (13 for Depth Anything and 17 for VGGT) and training on only about 300 annotated samples, 
we qualitatively observe strong generalization towards out of distribution 
indoor and outdoor ground types and that the network correctly omits flat surfaces such as walls and tabletops. 
We provide an example of ground segmentation in indoor and outdoor scenes in \cref{fig:ground_seg} using the VGGT head. 
Furthermore, to show our ground segmentation technique extends to larger GFMs, 
we qualitatively demonstrate accurate places reconstruction with Depth Anything
DA3NESTED-GIANT-LARGE-1.1 in \cref{sec:realtor} which has 40 layers, where layer 28 is optimal. 

\myParagraph{Forming the Places Layer}
Given ground masks, we get the corresponding labeled 3D points in the current submap and fit a plane to the points. 
On the plane, we bin the points into square tiles of size \tileSize, so that free space is stored as a sparse set of 
graph nodes rather than a dense occupancy grid. We remove tiles that do not contain points in all four of their quadrants, 
which filters out boundary tiles where the ground is only partially observed, and we remove tiles lying under points of 
vertical distance $d_{\text{max}}$, which avoids non-traversable areas such as underneath tables. To merge new tiles into existing 
tiles of prior submaps, we prune tiles with significant overlap and form new edges between tiles within 1.5\tileSize. 
During a loop closure, the places graph is deformable and tiles can be pruned and new edges formed as the tiles move.

The places graph thus forms a graph where nodes are the centroids of each tile and edges connect adjacent 
tiles that are traversable. We plan trajectories over the graph using Dijkstra. 

\subsection{Regions Layer}
\label{sec:regions}

Many tasks require reasoning about abstract region-level concepts rather than specific objects. For example, a task such as ``go to the kitchen'' requires understanding the spatial extent of the kitchen region. The boundary of a region is often ambiguous and can be defined at different granularities. For example a living room with a corner dedicated to kids' toys and activities, can be classified as a single room or into two regions (living room and playroom) depending on the the task.
This ambiguity is even more pronounced in outdoor environments where walls or furniture are not present to provide boundaries. Existing approaches~\cite{Hughes22rss-hydra,Hughes24ijrr-hydraFoundations,Werby24rss-hovsg} rely on indoor geometric priors such as detected walls and voxelized room boundaries that do not transfer to outdoor settings.

We sidestep this ambiguity by not committing to a global scene partition. Our regions layer is \emph{queryable}: given a natural-language query such as ``kitchen'' or ``parking lot'', it returns the set of places in the environment that belong to that region. A downstream task like ``bring me the towel in the kitchen'' is resolved against the region of places returned for the query ``kitchen'', so the towel in the bathroom is not a candidate answer.
Inspired by~\cite{Gorlo26cvpr-DAAAM,Hughes24ijrr-hydraFoundations,Strader24ral-automaticAbstractions}, the substrate for a region query is the places layer of our 3D scene graph. A region is therefore a subset of places nodes. 

\myParagraph{Place Semantic Statistic}
To cluster places nodes into regions, we need each place to carry semantic information.
Here, each place node summarizes the views from which it was observed into a single semantic statistic of Perception-Encoder (PE) embeddings. Na\"ively clustering based on averaged embeddings, as in DAAAM~\cite{Gorlo26cvpr-DAAAM}, fails on the places that are most important to disambiguation: a place at a region boundary sees both regions, and a place observed in views that mostly capture a wall carries an embedding that carries little semantic information. In both cases the cross-view variance is high and averaging is \emph{detrimental}~\cite{Kassab24arxiv-BareNecessities}. Storing every view per place would avoid this loss of information but scales poorly for clustering large scenes. 
We instead fit a von Mises-Fisher (vMF) distribution~\cite{Banerjee05jmlr-vMF} 
to the per-keyframe semantic embeddings that observe a place. This fixed-size summary captures both a mean direction $\boldsymbol{\mu}_i$ and a per-place concentration $\kappa_i$. The concentration is large for places observed many times from views that semantically agree with each other and small for boundary places and places with uninformative visual content.

\myParagraph{Query Scoring and Propagation}
An open-vocabulary region query is encoded via PE into a unit vector $\mathbf{q}$, and each place receives a match score $s_i = \kappa_i \langle \boldsymbol{\mu}_i, \mathbf{q} \rangle$, the cosine similarity between its mean view direction and the query, weighted by the place's observation confidence. Well-observed interior places contribute fully. Boundary and under-observed places, which a raw cosine match would most often mislabel, contribute proportionally less. To suppress residual single-view noise (\textit{e.g.,} an occlusion, viewing into the next room), we smooth the scores along the places graph with a confidence-aware variant of graph label propagation~\cite{Zhou03nips-lgc}: each place blends its own score with the average of its neighbors using a weight $\alpha_i = \kappa_i / (\kappa_i + \lambda \deg(i))$, where $\lambda$ is set from the median per-place concentration and $\deg$ is the node degree. Confident interior nodes stay close to their own score while under-observed boundary places inherit support from their neighborhood. Note that in this stage the topology of the places graph naturally limits propagation across narrow bottlenecks.

\myParagraph{Region Extraction}
After propagation, we separate places into in-region and out-of-region for the query by fitting a two-component one-dimensional Gaussian mixture to the propagated scores, and taking the component with higher mean as the in-region set. We then extract the connected components of in-region places on the places graph and return those above a minimum size. A query can therefore return multiple disjoint regions, \eg the two bedrooms of an apartment. 

\subsection{Agentic Comprehension}
\label{sec:agent}

Our representation can be used by an LLM-based agent to actively build a query-based 3D scene graph representation. In contrast to prior work using tool-calling agents for 3D scene understanding~\cite{Gorlo26cvpr-DAAAM,Anwar25icra-remembr}, our agent does not simply perform semantic similarity search and returns top-k results. Rather, by searching over relevant (semantically close) frames in the \emph{visual} memory and querying SAM3 within those frames as in VGGT-SLAM 2.0~\cite{Maggio26rss-VGGT-SLAM2}, the agent exhaustively searches for occurences of a queried object type in the scene and also understands when objects are \emph{not} present anywhere a scene.
The agent can query the scene graph for objects and regions relevant to a task. All queries get added to the cached memory, actively building up a scene graph representation directly relevant to the task. Based on these tools the agent can perform task-oriented reasoning as shown in \cref{sec:sg3d}.  %

\section{Experiments}
\label{sec:experiments}

In this section we run a suite of experiments demonstrating \name achieves superior performance on 
open-set 3D object extraction benchmarks (\cref{sec:clio}), more comprehensive 
object task-grounding with an agentic LLM (\cref{sec:sg3d}), and competitive room segmentation compared to methods using 
purely geometric clustering designed only for indoor scenes. 
Furthermore, we demonstrate \name's generality by running 
on in-the-wild internet videos (\cref{sec:realtor}) and in real-time onboard Spot (\cref{sec:spot}).

\subsection{Clio Open-Set 3D Object Extraction}
\label{sec:clio}

\begin{figure}[h]
    \centering
    \includegraphics[width=.48\textwidth]{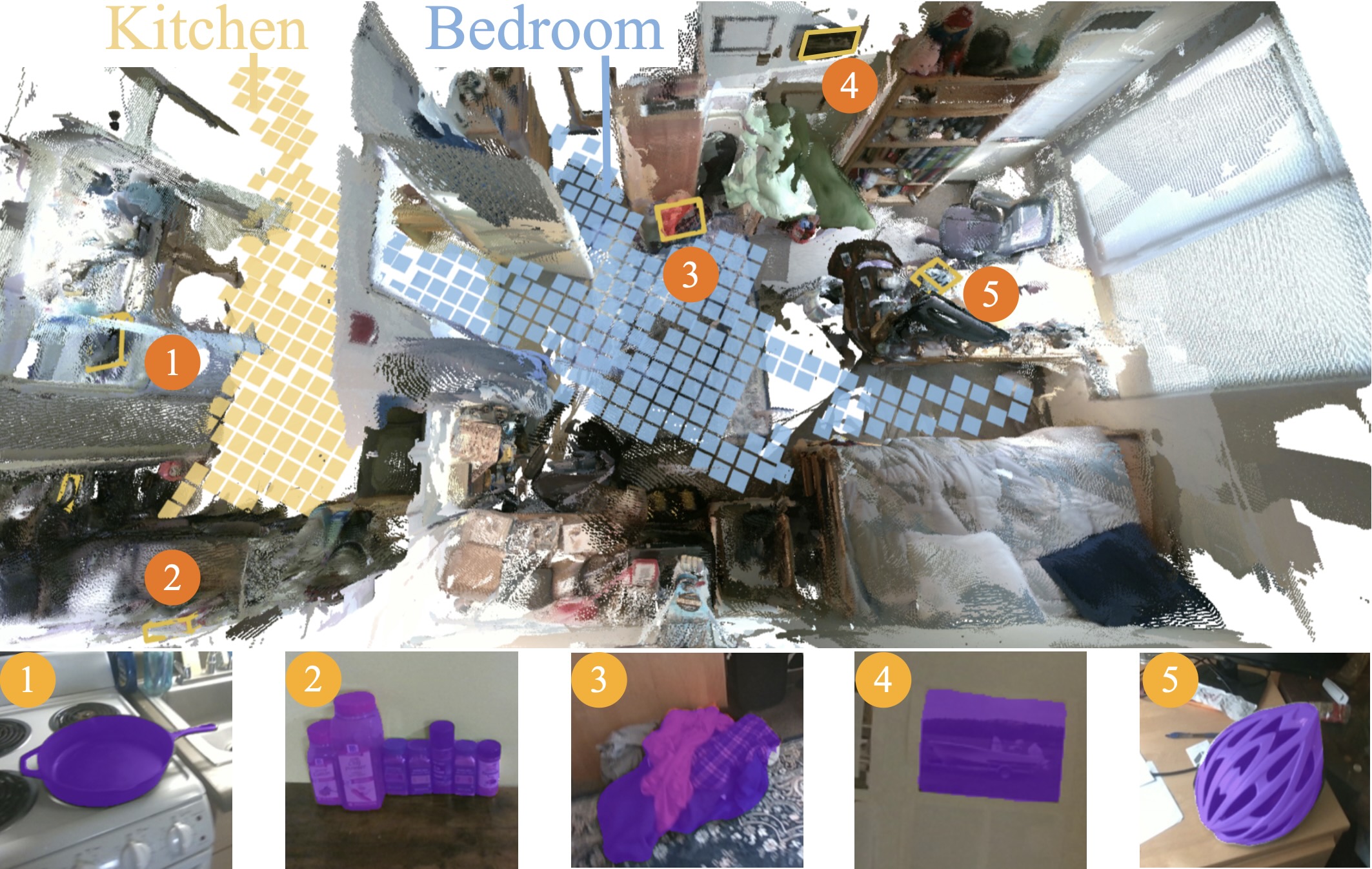}
    \caption{Scene reconstruction of the Clio~\cite{Maggio24ral-clio} apartment scene showing 2 queried rooms with their corresponding places tiles, 
    and 5 example queried objects with their 3D bounding box and corresponding segmented retrieved keyframe.}
    \label{fig:clio}
\end{figure}

\myParagraph{Setup}
We test open-set object extraction for all three scenes of the Clio dataset which consist of a cubicle, an apartment, and 
an office which contain 18, 28, and 33 objects respectively, using 
the same baselines as Bayesian Fields~\cite{Maggio25arxiv-BayesianFields}. 
We report two measures of accuracy being open-set recall (osR) 
which is the ratio of correctly mapped objects to total number of ground truth objects. An estimated object is considered 
correct if the 3D bounding box of the estimated object and 3D bounding box of the ground truth object both capture each other's 
centroid. Out of fairness, for quantitative evaluations on the Clio scenes we pass in the same ground truth poses and depth images 
to \name which are used by all methods. These take the place of the estimated depth and poses from the GFM and allow us to fairly 
compare each method's semantic extraction. To showcase our full pipeline, we include qualitative results using our 
default VGGT configuration in \cref{fig:clio}. 

Many of the Clio tasks are prompts that append the phrase ``get'' in front of the query object such as 
``get chapstick.'' This extra verbosity in the prompt generally works with CLIP 
since it tends to act like a bag of words~\cite{Yuksekgonul23iclr-clipBagOfWords} finder but is unsuitable for 
SAM3. Thus, we remove the extra verbosity and only use the name of the object. 
Furthermore, to make the correct granularity more explicit, %
we add context such as ``pile of'' when the object is to return a course granularity.

\myParagraph{Results}
In \cref{table:clio} we demonstrate \name achieves top performance on all three scenes of the Clio datasets, with 
substantial improvement in IoU, and is only method to have top performance across all scenes and metrics. We compare 
with both task-driven baselines (highlighted in blue) and non-task-driven open-set maps including Gaussian Splatting based 
methods (OpenGS, SemanticGS, and Bayesian Fields). Following~\cite{Maggio25arxiv-BayesianFields}, we omit 
OpenGS and SemanticGS on the office scene as their Gaussian Splat maps have poor reconstruction on office. 
While the task-driven baselines 
generally perform better than the task-agnostic baselines, the brittleness of the Information Bottleneck approach 
of Clio and Bayesian Fields can result in 
either over or under clustering which reduces IoU score. In contrast, our much simpler object visual memory captures the desired 
object with higher accuracy and reliability and unlike all other task-driven mapping systems, \emph{we do not rely on a predefined and fixed task list}. 

\begin{table}[h]
    \tiny
    \setlength{\tabcolsep}{4pt} %
    \centering
    \resizebox{\columnwidth}{!}{
    \begin{tabular}{l cccccc}
    \toprule
    \multirow{2}{*}{Method} & \multicolumn{2}{c}{Cubicle} & \multicolumn{2}{c}{Apartment} & \multicolumn{2}{c}{Office} \\
    \cmidrule(lr){2-3} \cmidrule(lr){4-5} \cmidrule(lr){6-7}
    & osR$\uparrow$ & IoU$\uparrow$ & osR$\uparrow$ & IoU$\uparrow$ & osR$\uparrow$ & IoU$\uparrow$ \\
    \midrule
    CG~\cite{Gu24icra-conceptgraphs} & 0.44 & 0.06 & 0.38 & 0.07 & 0.24 & 0.07 \\ 
    Khronos~\cite{Schmid24rss-khronos} & 0.78 & 0.17 & 0.45 & 0.11 & \underline{0.67} & \underline{0.15} \\
    Clio-Prim~\cite{Maggio24ral-clio} & 0.72 & 0.18 & 0.35 & 0.12 & \textbf{0.70} & 0.17 \\
    \sg~\cite{Guo24arxiv-semanticGS} & 0.17 & 0.03 & 0.00 & 0.00 & -- & -- \\
    \ogs~\cite{Wu24neurips-opengaussian} & 0.78 & 0.07 & 0.31 & 0.06 & -- & -- \\
    \rowcolor{blue!20} CG-\thres & 0.44 & 0.06 & 0.21 & 0.03 & 0.19 & 0.06 \\
    \rowcolor{blue!20} Khronos-\thres & 0.78 & 0.17 & 0.41 & 0.11 & 0.55 & 0.12 \\
    \rowcolor{blue!20} Clio-\batch~\cite{Maggio24ral-clio} & \underline{0.83} & 0.17 & 0.52 & 0.11 & 0.64 & 0.13 \\
    \rowcolor{blue!20} Clio-online~\cite{Maggio24ral-clio} & \textbf{0.89} & 0.22 & 0.35 & 0.07 & 0.55 & 0.12 \\
    \rowcolor{blue!20} Bayesian Fields~\cite{Maggio25arxiv-BayesianFields} & \textbf{0.89} & \underline{0.26} & \underline{0.59} & \underline{0.19} & 0.55 & \underline{0.17} \\
    \rowcolor{blue!20} \textbf{\name (Ours)} & \textbf{0.89} & \textbf{0.37} & \textbf{0.66} & \textbf{0.28} & \textbf{0.70} & \textbf{0.30} \\
    \bottomrule
    \end{tabular}
    }
    \caption{Evaluation of open-set 3D object extraction on the Clio~\cite{Maggio24ral-clio} datasets. 
    Methods highlighted in blue are task-driven.}
    \label{table:clio}
\end{table}

To demonstrate our full system running on the Clio scenes (including VGGT estimated poses), 
we show our 3D scene graph --- including example extracted objects, the 
places layer, and example regions in \cref{fig:clio} for the apartment scene. 

\subsection{SG3D Task Grounding}
\label{sec:sg3d}

\myParagraph{Setup}
We evaluate our performance in task grounding following the protocol of ASHiTA~\cite{Chang25cvpr-ashita} and DAAAM~\cite{Gorlo26cvpr-DAAAM} on the SG3D~\cite{Zhang24arxiv-taskOrientedGrounding} benchmark. The benchmark evaluates the ability of a 3D scene representation to support task grounding: grounding the location of a target object instance given a task description. The benchmark uses eight HM3D-SEM validation scenes, and evaluates on 1000 queries per scene, where each query consists of a target object instance and a task description. The task description is a natural language instruction describing the task to be performed with the target object, and may include references to other objects in the scene. The evaluation metric is the success rate of grounding the target object instance within a certain distance threshold.
Our LLM agent, described in~\cref{sec:agent} can query our scene graph representation for objects and return the final location of the target object instance. In order to match ASHiTA~\cite{Chang25cvpr-ashita}, we use GPT-4o-mini for all methods. 

\myParagraph{Results}
\cref{tab:grounding_results} shows our results compared to ASHiTA~\cite{Chang25cvpr-ashita}, DAAAM~\cite{Gorlo26cvpr-DAAAM}, and two baselines which use the Hydra~\cite{Hughes24ijrr-hydraFoundations} and HOV-SG~\cite{Werby24rss-hovsg} region representations as input to GPT-4o-mini. Our method outperforms all baselines by a significant margin, demonstrating the effectiveness of our queryable representation for task grounding.
Importantly, as our memory representation inherits the concept of \emph{nonentity} from SAM3, our agent can understand when an object is \emph{not} present anywhere in the scene and query the scene graph for a different class of objects that may be relevant to the task.

\setlength{\tabcolsep}{4pt} %
\begin{table}[h]
\vspace{-5pt}
\centering
\adjustbox{width=.95\columnwidth}{%
\scriptsize
\begin{tabular}{lcc}
    \toprule
    Method & s-acc [\%] $\uparrow$ & t-acc [\%] $\uparrow$ \\
    \midrule
    Hydra~\cite{Hughes24ijrr-hydraFoundations} + GPT-4o-mini  & 8.18 & 2.44 \\
    Hydra~\cite{Hughes24ijrr-hydraFoundations} (GT Seg) + GPT-4o-mini & 14.2 & 6.34 \\
    HOV-SG~\cite{Werby24rss-hovsg} & 8.98 & 1.95  \\
    ASHiTA~\cite{Chang25cvpr-ashita} & 21.7 & 8.78 \\
    DAAAM~\cite{Gorlo26cvpr-DAAAM} + GPT-4o-mini & \underline{22.2} & \underline{11.2} \\
    \textbf{\name} (Ours) + GPT-4o-mini & \textbf{39.7} & \textbf{19.0} \\
    \bottomrule
\end{tabular}}
\caption{Results on sequential task grounding for the ASHiTA SG3D~\cite{Zhang24arxiv-taskOrientedGrounding} benchmark. All methods use depth data and ground-truth poses from the benchmark dataset.}\label{tab:grounding_results}
\end{table}

\subsection{Region Clustering}
\label{sec:exp_regions}

\myParagraph{Benchmark and metrics}
We evaluate our open-vocabulary regions layer on the HOV-SG~\cite{Werby24rss-hovsg} room-segmentation benchmark, which uses $8$ multi-floor HM3D-SEM validation scenes and a fixed closed vocabulary of indoor room categories. Following HOV-SG, we report three metrics: region precision and recall on the 2D floor plane, and the semantic accuracy $\text{Acc}_{=}$ of the predicted region label against the ground-truth category, evaluated given the ground-truth region segmentation.

\myParagraph{Adapting to the Closed-vocabulary Dense Benchmark}
Our regions layer (\cref{sec:regions}) is designed for open-vocabulary, query-driven retrieval. At runtime, an agent issues a free-text query and the layer returns the matching places clusters as regions (between none and several disjoint regions). HOV-SG, like prior room-segmentation methods~\cite{Werby24rss-hovsg,Hughes24ijrr-hydraFoundations,Bormann16icra-roomSegmentationSurvey}, instead expects three things our method does not natively produce: (a) a single scene partition over a fixed closed vocabulary of indoor rooms, (b) dense coverage of the wall-bounded 2D ground-truth mask (subject to indoor structural priors) even on cells our sparse traversability graph does not reach (\eg under furniture), and (c) a prediction emitted for every category. We address each mismatch with a single adaptation, leaving the per-place vMF statistic, scoring, and graph propagation of \cref{sec:regions} unchanged.

First, for the global partition (a), we run the per-query scorer once per category and normalize each query's scores so that a global CLIP attractor (\eg ``entryway'') does not supress the others, and assign each place to its best scoring category.
Second, for dense coverage (b), we BFS-propagate labels along the traversability graph until we have full coverage.
Third, for within-category fragmentation and the requirement of always returning a corresponding region (c), we run Louvain~\cite{Blondel08jsm-Louvain} on the places graph with edges weighted by how similarly adjacent places score the closed-set categories, and relabel each community by its highest scoring category. This combines a fragmented room with the same category back into one community and lets every category claim at least one connected community.

\myParagraph{Results}
Table~\ref{tab:hovsg_regions} compares our method to HOV-SG~\cite{Werby24rss-hovsg} and Hydra~\cite{Hughes24ijrr-hydraFoundations} on the 8 validation scenes. Our precision and recall are competitive despite a representation that relies less on indoor priors. On semantic accuracy $\text{Acc}_{=}$, we exceed HOV-SG by 4.12 points. HOV-SG aggregates embeddings of frames of which the camera pose falls in a room, so a camera in the kitchen looking into the living room pollutes the kitchen with a living-room view. Our per-place vMF statistic (\cref{sec:regions}) instead attaches views to the place being observed.

\begin{table}[h]
    \tiny
    \setlength{\tabcolsep}{4pt} %
    \centering
    \resizebox{\columnwidth}{!}{
    \begin{tabular}{l ccc}
    \toprule
    Method & Precision$ [\%]\uparrow$ & Recall$ [\%]\uparrow$ & $\text{Acc}_{=}$$ [\%]\uparrow$ \\
    \midrule
    Hydra~\cite{Hughes24ijrr-hydraFoundations} & \textbf{86.18} & 77.55 & -- \\
    HOV-SG~\cite{Werby24rss-hovsg} & 84.10 & \underline{83.59} & \underline{73.93} \\
    GPT-4 w/ HOV-SG & \multirow{2}{*}{--} & \multirow{2}{*}{--} & \multirow{2}{*}{59.47} \\
    object categories & & & \\
    \textbf{\name (Ours)} & \underline{85.34} & \textbf{83.72} & \textbf{78.05} \\
    \bottomrule
    \end{tabular}
    }
    \caption{Evaluation of region clustering and labeling on the HOV-SG~\cite{Werby24rss-hovsg} benchmark. Best is bold, second underlined. Our method is competitive with state-of-the-art methods despite relying less on indoor priors.}
    \label{tab:hovsg_regions}
\end{table} 
\subsection{In-the-wild Internet Scenes}
\label{sec:realtor}
To demonstrate \name can be easily run on in-the-wild scenes, we demonstrate mapping on cell phone videos of realtor home tours 
from YouTube\footnote{Used with permission from raw video files provided by the realtor.}. In \cref{fig:intro} we show 
example room and object extraction on two scenes, with additional scenes (including a multi-story home) provided 
in the supplementary video\footnote{\url{https://www.youtube.com/watch?v=gaPeSTlYQKE}} for a total of five scenes. Due to challenging camera motions, 
we obtain the best geometric performance 
using the larger Depth Anything DA3NESTED-GIANT-LARGE-1.1 model, which we easily plug into our \name pipeline as the GFM. 

\subsection{Additional Qualitative results}
\label{sec:qualitative}

\begin{figure}[h]
    \centering
    \vspace{0.5mm}
    \begin{subfigure}[b]{0.47\textwidth}
        \includegraphics[width=\textwidth]{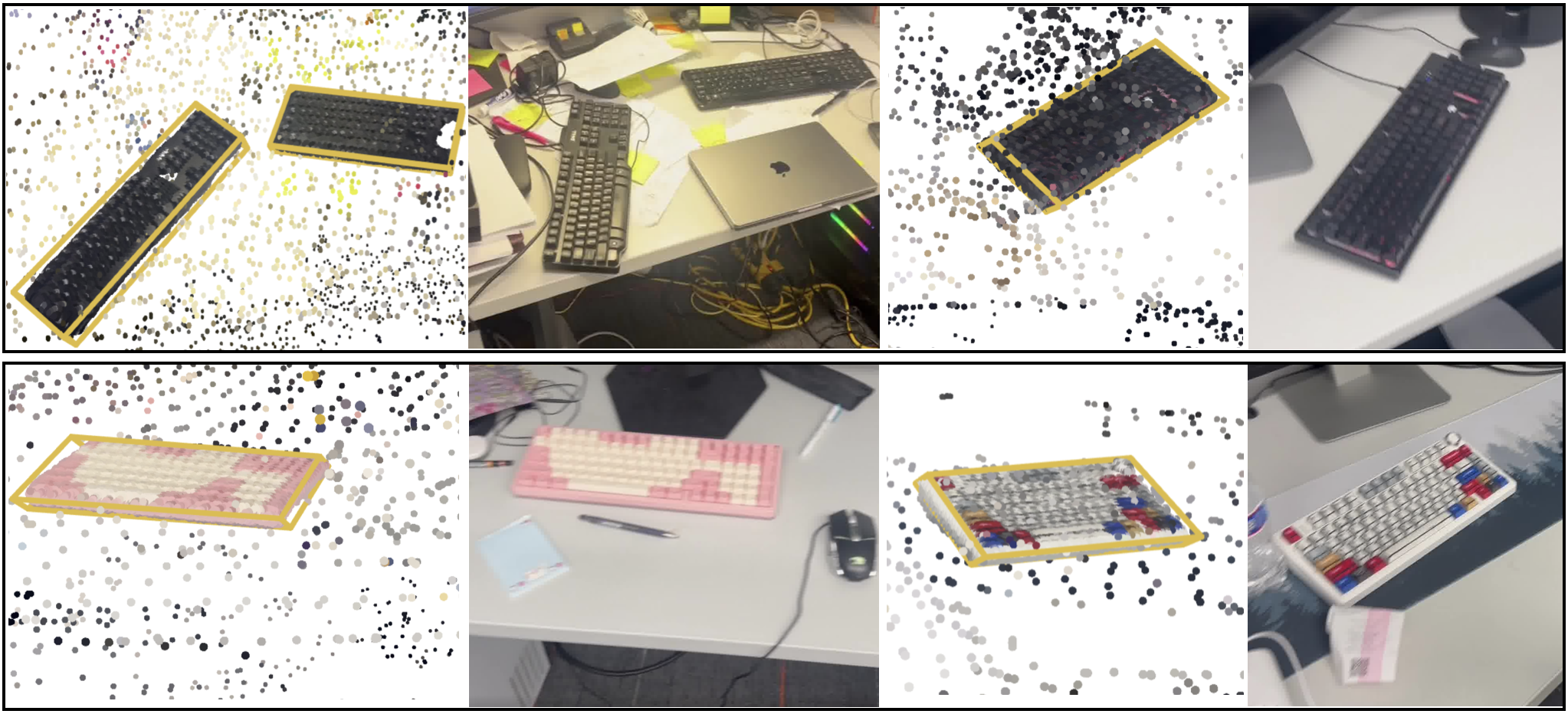}
        \caption{Example of finding multiple object instances for the same query (in this case 5 keyboards for the query ``keyboard'')}
        \label{fig:keyboards}
    \end{subfigure}
    \hspace{\fill}
    \begin{subfigure}[b]{0.47\textwidth}
        \includegraphics[width=\textwidth]{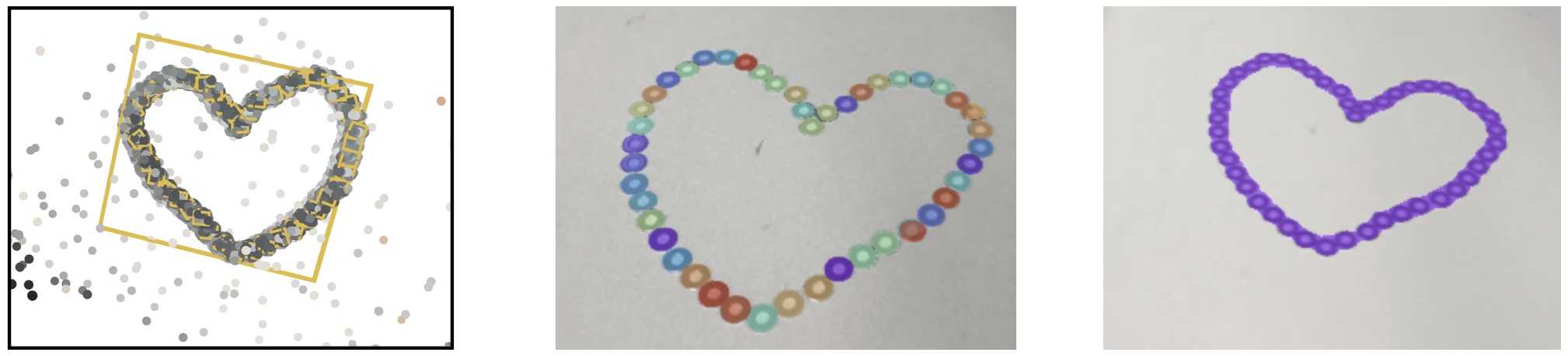}
        \caption{Example of using queries at two different granularities (``heart'' and ``washer'') and demonstrating detection 
        of many objects (45 washers). 
        Left: 3D bounding boxes for both queries. Middle and Right: Corresponding SAM3 segmented keyframe for each query.}
        \label{fig:heart}
    \end{subfigure}
    \caption{Example of retrieving multiple objects from one query and representing a scene with different task-driven granularities. 
    We also demonstrate keeping the overall scene sparse while using dense points for queried objects.}
    \label{fig:sparse}
\end{figure}

In \cref{fig:sparse} we provide an example of generating a sparse map while only storing dense 3D points for queried objects using the 
method described in \cref{sec:objects}. 
Furthermore, in \cref{fig:keyboards} we demonstrate detection of multiple object instances for one query, in this case 5 keyboards which 
required retrieving four keyframes from visual memory for 3D object detection. 
In \cref{fig:heart} we provide an example of multi-granularity 
task-driven querying where, given a heart made of washers, our map can both represent the 
region as one object (the heart) or as 45 objects (each washer). 

\begin{figure}[h]
    \centering
    \includegraphics[width=.48\textwidth]{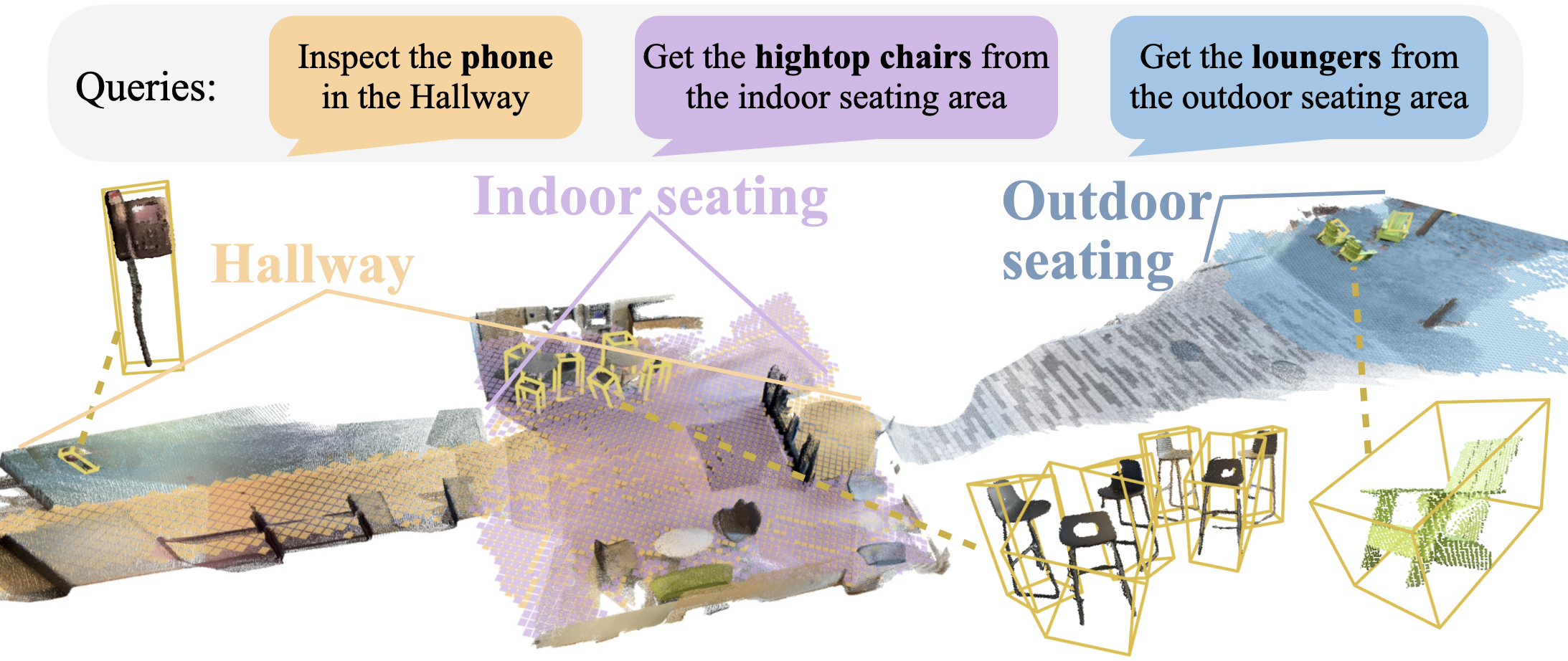}
    \vspace{-2mm}
    \caption{An LLM agent orchestrates \name to process the given tasks and form a 3D scene graph spanning indoor and outdoor areas.
    The object layer uses a visual memory which returns explicit 3D objects during open-set querying (here we show objects for ``phone'', ``hightop chairs'', ``green loungers''). The places layer maps 
    traversable space that is clustered into task-relevant regions at query-time. The system can determine regions at multiple granularities, \eg ``hallway'' and ``indoor seating'' (a subset of the hallway).}
    \label{fig:indoorOutdoor}
\end{figure}

In \cref{fig:indoorOutdoor} we show an LLM-agent using \name to construct a 3D scene graph in indoor and outdoor environments by grounding three tasks. We provide additional examples of \name constructing scene graphs in outdoor environments and querying objects at multiple granularities in our supplementary video.

\subsection{Timing and Memory Usage}
\label{sec:timing}

Using a 3090 GPU, and the Clio apartment scene as an example dataset, we report 
timing for the primary components of \name in \cref{tab:timing}, demonstrating real-time performance. We 
report the total per-submap time for submaps of size 16 keyframes and overall fps as total time per submap divided by submap size. 
Using VGGT, our entire pipeline runs at 6 fps, with the GFM being the slowest component. 
Using the lighter-weight Depth Anything DA3-LARGE-1.1 model, our pipeline runs at 7 fps. 
The time to query a single object from visual memory is approximately 100ms, and the time to cluster and 
extract a region is 530ms. 

\begin{table}[hbt]
    \centering
    \vspace{0.5mm}
    \adjustbox{width=.95\columnwidth}{
    \scriptsize
    \begin{tabular}{l|cc}
        \toprule
        Component & Total Time per Submap (ms) \\
        \hline
        VGGT Inference & 1333 ms \\
        Geometric Mapping (excluding VGGT) & 159 ms \\
        Object Visual Memory Layer & 393 ms \\
        Places and Regions & 140 ms \\
        \bottomrule
    \end{tabular}}
    \caption{Average runtime per submap in milliseconds for the primary components of \name using a 
     3090 GPU with submap size of 16 frames. 
     Geometric mapping time includes time for keyframe detection, image retrieval, and factor graph optimization.}\label{tab:timing}
    \end{table}

\subsection{Real-time Test Onboard Spot}
\label{sec:spot}
We demonstrate our full 3D scene graph can be constructed in real-time using a Jetson Thor mounted on a Spot quadruped robot. 
Since the most computationally extensive component of \name is the GFM, we can leverage our ability to 
easily exchange GFMs by using the lighter-weight Depth Anything DA3-LARGE-1.1 model, enabling our full pipeline 
to run at 4 Hz on the Thor. 
In our supplementary video, we 
demonstrate open-set object querying and region querying as the robot maps the scene in real-time. %

\section{Limitations}
\label{sec:limitations}
While our system substantially improves upon task-driven, open-set, 3D scene graph construction, it is not without limitations. 
For instance, while we observe resiliency in our object detection, a failure in the geometric mapping system 
will lead to failure in the scene graph construction. To reduce this limitation, we demonstrate 
flexibility in the mapping system by showing \name can use 
multiple geometric foundation models. We also do not address dynamic objects in this work, 
although \name's object cache memory could be used to capture changes in an object's location.
While our places and region representation provides regions at the desired granularity to ground robot instructions, the traversability estimation (and thereby the region clustering) 
assumes there are some keyframes which observe the floor.  %

\section{Conclusion}
\label{sec:conclusion}

We have presented \name, the first system for constructing real-time, task-driven, hierarchical, open-set 3D scene graph built on top of 
geometric foundation models. 
By using a visual memory representation, we are able to query dense 3D point clouds of objects at the 
correct task-driven granularity to support the agent's tasks. By leveraging the visual embeddings of the  
geometric foundation model to predict ground segmentation, we are able to seamlessly generate traversable 
places of the scene graph and can cluster these places into regions which are defined quickly at query time to 
maintain task-relevant querying. We have also demonstrated the universality of our system by running a range of 
indoor and outdoor scenes and on casually captured YouTube videos. 
\section*{Acknowledgement}
We gratefully thank BBA Management for providing us with apartment tour videos for experimental evaluation. 

{
\bibliographystyle{IEEEtran}

}

\end{document}